\documentclass[preprint,3p]{elsarticle}

\journal{Journal of Biomedical Informatics}

\usepackage{amsmath}
\usepackage{booktabs}
\usepackage{adjustbox}
\usepackage{makecell}
\usepackage{float}
\usepackage{subcaption}
\usepackage{multirow}
\begin{document}

\begin{frontmatter}

\title{Patient representation learning and interpretable evaluation using clinical notes}


\author[uzainf,clips]{Madhumita Sushil\corref{mycorrespondingauthor}}
\cortext[mycorrespondingauthor]{Corresponding author}
\ead{madhumita.sushil@outlook.com}
\author[clips]{Simon \v{S}uster}
\author[uzainf]{Kim Luyckx}
\author[clips]{Walter Daelemans}

\address[uzainf]{Antwerp University Hospital, ICT department, Wilrijkstraat 10, Edegem, 2650 Belgium}
\address[clips]{Computational Linguistics and Psycholinguistics (CLiPS) research center, University of Antwerp, Prinsstraat 13, Antwerp, 2000 Belgium}

\begin{abstract}
We have three contributions in this work: 
1. We explore the utility of a stacked denoising autoencoder and a paragraph vector model to learn task-independent dense patient representations directly from clinical notes. To analyze if these representations are transferable across tasks, we evaluate them in multiple supervised setups to predict patient mortality, primary diagnostic and procedural category, and gender. We compare their performance with sparse representations obtained from a bag-of-words model. We observe that the learned generalized representations significantly outperform the sparse representations when we have few positive instances to learn from, and there is an absence of strong lexical features.
2. We compare the model performance of the feature set constructed from a bag of words to that obtained from medical concepts. In the latter case, concepts represent problems, treatments, and tests. We find that concept identification does not improve the classification performance.
3. We propose novel techniques to facilitate model interpretability. To understand and interpret the representations, we explore the best encoded features within the patient representations obtained from the autoencoder model. Further, we calculate feature sensitivity across two networks to identify the most significant input features for different classification tasks when we use these pretrained representations as the supervised input. We successfully extract the most influential features for the pipeline using this technique.
\end{abstract}


\end{frontmatter}


\section{Introduction}
\label{sec:intro}

Representation learning refers to learning features of data that can be used by machine learning algorithms for different tasks. Sparse representations, such as a bag of words from textual documents, treat every dimension independently. For example, in one-hot sparse representations, the terms `pain' and `ache' correspond to separate dimensions despite being synonyms of each other. Several techniques exist to model such dependence and reduce sparsity. The generalized or distributed representations learned using these techniques are referred to as low dimensional, or dense data representations. Unsupervised techniques for representation learning have become popular due to their ability to transfer the knowledge from large unlabeled corpora to the tasks with smaller labeled datasets, which can help circumvent the problem of overfitting~\citep{goodfellow2016deep}. 

Representation learning techniques have been used extensively within and outside the clinical domain to learn the semantics of words, phrases, and documents~\citep{DBLP:conf/acl/BaroniDK14,DBLP:journals/corr/LiuCJY16}. We apply such techniques to create a patient semantic space by learning dense vector representations at the patient level. In a patient semantic space, ``similar" patients should have similar vectors. Patient similarity metrics are widely used in several applications to assist clinical staff. Some examples are finding similar patients for rare diseases~\citep{garcelon2017finding}, identification of patient cohorts for disease subgroups~\citep{li2015identification}, providing personalized treatments~\citep{zhang2014towards,wang2015electronic}, and predictive modeling tasks such as patient prognosis~\citep{gottlieb2013method,wang2012medical} and risk factor identification~\citep{ng2015personalized}. The notion of patient similarity is defined differently for different use cases. When it is defined as an ontology-guided distance between specific structured properties of patients such as diseases and treatments, it represents patient relationships corresponding to those properties. For example, if patient similarity is calculated as a hierarchical distance between the primary diagnostic codes of patients in the UMLS\textsuperscript{\textregistered} metathesaurus~\citep{LindbergEtAl1993}, the value represents a diagnostic similarity. When it is defined as an intersection between the sets of blood tests performed on patients, patient similarity maps to blood test similarity. If patient similarity value is 1 for the patients of the same gender and 0 otherwise, groups of similar patients are gender-specific patient cohorts. However, when we calculate similarity between distributed patient representations, the different properties that influence the similarity value are unknown. Within the learned patient representations, we aim to capture similarity on multiple dimensions, such as complaints, diagnoses, procedures performed, etc., which would encapsulate a holistic view of the patients. 

In this work, we create unsupervised dense patient representations from clinical notes in the freely available MIMIC-III database~\citep{johnson2016mimic}. We aim to learn patient representations that can later be used to identify sets of similar patients based on representation similarity. We focus on different techniques to learn patient representations using only textual data. We explore the usage of two neural representation learning architectures---a stacked denoising autoencoder~\citep{vincent2010stacked}, and a paragraph vector architecture ~\citep{le2014distributed}---for unsupervised learning. We then transfer the representations learned from the complete patient space to different supervised tasks, with an aim to generalize better on the tasks for which we have limited labeled data.

Dense representations can capture semantics, but at a loss of interpretability. Yet, it is critical to understand model behavior when statistical outputs influence clinical decisions~\citep{caruana2015intelligible}. We take a step towards bridging this gap by proposing different techniques to interpret the information encoded in the patient vectors, and to extract the features that most influence the classification output when these representations are used as the input.

\section{Related Work}
\label{sec:relwork}

\textit{Dense representations} of words ~\citep{mikolov2013efficient,mikolov2013distributed,pennington2014glove,DBLP:journals/tacl/BojanowskiGJM17} and documents~\citep{le2014distributed,larochelle2012neural} have become popular because they are learned using unsupervised techniques, they capture the semantics in the content, and they generalize well across multiple tasks and domains. An \textit{autoencoder} learns the data distribution and the corresponding dense representations in the process of first encoding data into an intermediate form and then decoding it. \citet{miotto2016deep} first proposed the use of a stacked denoising autoencoder to learn patient representations. They have shown promising results when patient vectors are first learned by a stacked denoising autoencoder from structured data combined with 300 topics from unstructured data, and are then used with Random Forests classifiers to identify future disease categories of patients. 
Following their work, \citet{dubois2017learning} have proposed two techniques to obtain patient representations from clinical notes. The first technique is unsupervised and performs an aggregation of concept embeddings into note and patient level representations, known as `embed-and-aggregate'. The second technique uses a recurrent neural network (RNN) with a bag-of-concepts representation of patient notes as time steps. The RNN is trained to predict disease categories of patients. The representations learned in this supervised setup are then transferred to other tasks. Apart from these works, \citet{suresh2016use} have performed a preliminary exploration of the use of sequence-to-sequence autoencoders to induce patient phenotypes using structured time-series data. They have compared different autoencoder architectures based on their reconstruction error when they are trained to encode patient phenotypes. An application of these phenotypes to different clinical prediction tasks has been reserved for future work. In the same vein as these previous works, we investigate the applicability of a stacked denoising autoencoder to learn patient representations \textit{directly from unstructured data}, and analyze the tasks that these representations can be successfully applied to.


One of the evaluation tasks for us is \textit{patient mortality prediction}. \citet{johnsonreproducibility} provide a good overview of the previous approaches for mortality prediction on the MIMIC datasets with an aim of replicating the experiments. Following the work by \citet{ghassemi2014unfolding}, \citet{grnarova2016neural} have shown significant improvements for mortality prediction tasks on using a two-level convolutional neural network (CNN) architecture, as compared to the use of topic models and doc2vec representations as inputs to linear support vector machines (SVMs). Besides these works, \citet{jocombining} have recently used long short term memory networks (LSTMs) and topic modeling for mortality prediction. They treat topics for patient notes as time steps for LSTMs. These topics are learned jointly using an encoder network. They have shown performance gains when the topics are jointly learned, compared to those pretrained using LDA~\citep{blei2003latent}. 

\section{Methods}
\label{sec:methods}

\subsection{Learning Patient Representations}

\begin{figure}
\centering
\includegraphics[width=0.5\linewidth]{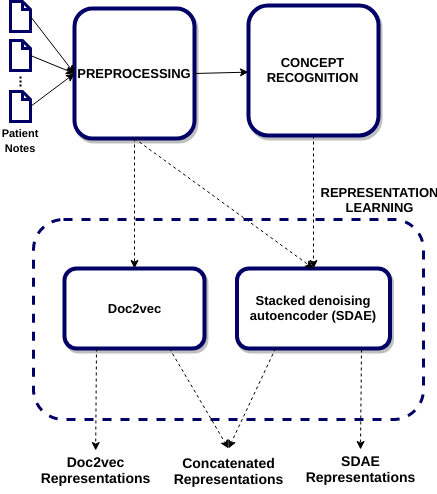}
\caption{An overview of the patient representation pipeline. The dashed lines indicate one of several operations, and are not performed in parallel.
}
\label{fig:patient_rep}
\end{figure}

In this section, we describe a stacked denoising autoencoder and a paragraph vector architecture doc2vec, in the context of learning task-independent dense patient representations in an unsupervised manner. The corresponding methodology for learning these dense representations is illustrated in Figure~\ref{fig:patient_rep}.

\subsubsection{Stacked denoising autoencoder}
\label{sec:sdae}

Given the previous success of autoencoders for representation learning using structured data with or without topic models learned from unstructured data, we explore the use of a stacked denoising autoencoder (SDAE)~\citep{vincent2010stacked} to learn task-independent patient representations from raw clinical text, forgoing the use of intermediate techniques like topic modeling. Although the premise of learning patient representations using an SDAE is not novel in itself, our contribution lies in analyzing if such a model is also successful when used only with clinical notes, and if the learned representations can be successfully applied for a range of tasks that are different from patient prognosis. This analysis gives us insight into successful and transferable patient representation architectures for unstructured data.

During the \textbf{pretraining} phase, every layer of an SDAE is sequentially trained as an independent denoising autoencoder. An autoencoder learns to first encode the input data $I$ into an intermediate representation $R$, and then decode $R$ into $I$. Denoising refers to the process of first adding noise to corrupt the input $I$ into $\widetilde{I}$, and then training an autoencoder to reconstruct $I$ using $\widetilde{I}$ as the input. We use the dropout noise~\citep{srivastava2014dropout}, where a random proportion of the input nodes are set to 0. In the process of denoising, the model also learns the data distribution. In an SDAE, the intermediate representations obtained from the autoencoder at layer $n-1$ are used as the uncorrupted input to the autoencoder at layer $n$, for all the layers in the SDAE. To pretrain patient representations using an SDAE, high-dimensional (sparse) patient data are used as the input to the autoencoder at the first layer of the SDAE. The intermediate representations obtained from the autoencoder at the final layer are treated as the low-dimensional (dense) representations $R(p)$ for a patient $p$. The number of layers is determined through a random search~\citep{bergstra2012random} based on the results for primary diagnostic category prediction using a perceptron.

\textbf{Finetuning} can be performed in multiple ways~\citep{goodfellow2016deep}. In one approach, all the encoder layers can be stacked on top of each other, and a logistic regression layer can be added on the top to finetune the entire pretrained network for an end task as a feedforward neural network. In such a setup, the input features in the finetuning phase are the same as the input features during the pretraining phase. In another approach, instead of the entire network, only the preliminary task-independent representations $R$ can be finetuned for an end task. In this approach, $R$ is used as the input to a separate classifier. In our experiments, we train separate classifiers for different tasks using $R$ as the input features.

We use the sigmoid activation function for the encoding layers, and the linear activation function to decode real values. During the pretraining phase, we train each layer of the SDAE to minimize the mean squared reconstruction error using the RMSProp optimizer~\citep{tieleman2012lecture}. During the finetuning phase, we train the classifiers to minimize the categorical cross-entropy error using the same optimizer. We determine the number of layers, the dimensionality, and the dropout proportion also using a randomized hyperparameter search. These values are dependent on the feature sets and the finetuning process, and can be found in Table~\ref{hyperparams-sdae} in the Appendix.

\subsubsection{Paragraph vector}
\label{sec:doc2vec}
\textbf{Doc2vec}, or `Paragraph Vector'~\citep{le2014distributed}, learns dense fixed-length representations of variable length texts such as paragraphs and documents. It supports two algorithms---a distributed bag-of-words (DBOW) algorithm, and a distributed memory (DM) algorithm. For both the algorithms, word representations are shared among all the occurrences of a word across all the paragraphs, and paragraph vectors are shared among all the contexts that occur in a given paragraph. In the DBOW algorithm, word and paragraph vectors are jointly trained when the paragraph vectors are used to predict the context words for all the contexts in the paragraph. In the DM algorithm, these vectors are jointly trained by predicting the next word from a concatenation of the paragraph vectors and the vectors of the context words. During the inference phase of both the algorithms, word vectors are fixed, and paragraph vectors are trained until convergence. 

We use the DBOW algorithm for 5 iterations, with a window size of 3, a minimum frequency threshold of 10, and 5 negative samples per positive sample to train 300-dimensional patient vectors. We determined these settings also using randomized hyperparameter search.

\subsection{Feature extraction}

When statistical models are deployed for clinical decision support, it is crucial to understand the features that influence the model output~\citep{caruana2015intelligible}. A ranked list of the most influential features can assist such understanding, while facilitating error analysis; it can also enable exploratory analysis when unexpected features are ranked high. However, neural networks are notorious for being black boxes due to their complex architectures. Given the impact of automated decisions, there has been a recent surge of interest to make neural architectures interpretable. Different techniques include visualization of weights and embeddings~\citep{visualization_techreport, DBLP:conf/naacl/LiCHJ16}, representation erasure and feature occlusion~\citep{DBLP:journals/corr/LiMJ16a, DBLP:journals/corr/SureshHJCSG17}, input perturbation~\citep{DBLP:conf/emnlp/Alvarez-MelisJ17}, and visualization of attention weights in recurrent neural networks~\citep{DBLP:journals/corr/BahdanauCB14,DBLP:conf/nips/HermannKGEKSB15,DBLP:conf/naacl/YangYDHSH16, choi2016retain}. The technique of visualizing hidden weights and embeddings is a qualitative approach to interpretability. Furthermore, techniques like input feature erasure train a new model in absence of a given feature. When retrained, these models can learn to rely on a completely different set of features. Moreover, the attention mechanism is not applicable to feedforward neural networks. Within the scope of our work, we propose two techniques to bridge the existing gap in model interpretability when we train unsupervised dense representations, and when we use these representations to get classification decisions using feedforward neural networks. To the best of our knowledge, we are the first to propose these techniques to make dense representations interpretable.

\subsubsection{Average feature reconstruction error: pretraining phase}

We calculate the \textbf{squared reconstruction error} of all the input features in the first layer of the pretrained autoencoder, averaged across all the training instances. The value of the reconstruction error of the individual features gives us an estimate of the features that are encoded the best and the worst in the patient vectors learned through the SDAE. This knowledge facilitates an analysis of model behavior to make the vectors more interpretable.

\subsubsection{Input significance calculation using sensitivity analysis: classification phase}

\textbf{Sensitivity analysis}, or gradient-based analysis, is often used to identify the most influential features of a trained model~\citep{engelbrecht1998feature, dimopoulos1995use, gevrey2003review}. For a given model and a given instance, the sensitivity of an output node with respect to an input node refers to the observed variation in the output on varying the input. This is equivalent to the gradient of the output with respect to the input. The inputs that cause larger variations in the output are more significant for the model.

This analysis has so far been used to identify the most influential features for a single network, such as a single classifier. However, in our work, we are confronted with two neural networks. The first network learns the dense patient representations, and the second network uses these dense representations as the input for different classification tasks. We extend the work by~\citet{engelbrecht1998feature} and propose a technique to compute the significance of the original (sparse) features on the final classification decisions. We use the chain rule across two networks to compute the sensitivity of the output node in the second network to the input of the first network. This allows us to identify the most influential features in the entire pipeline. 

We demonstrate this technique for different classification tasks when the task-independent dense patient representations $R$ are first induced by the SDAE from the original input $z$, and $R$ is then used as the input to the classifiers. The significance of the $i$th input feature ($\phi_{z_i}$) is defined as the maximum significance of the input feature $i$ across all the $K$ output units ($o$) of the classifier with respect to the $N$ instances:

\begin{align}
\phi_{z_i} = \max_{k = 1...K}\{S_{o_k z_i}\} 
&\text{,\;where\;}
\end{align}

\begin{equation}
S_{o_k z_i} = \sqrt{\frac{\sum\limits_{j=1}^N [S_{o_k z_i}^{(j)}]^2 }{N}}. 
\end{equation}

\noindent $S_{o_k z_i}^{(j)}$ is the sensitivity of the $k$th output unit of the classifier w.r.t the $i$th input feature of the SDAE for an instance $j$: 

\begin{equation}
S_{oz,ki}^{(j)} = \frac{\partial o_k^{(j)}}{\partial z_i^{(j)}} =   \frac{\partial o_k^{(j)}}{\partial R_i^{(j)}} *  \frac{\partial R_i^{(j)}}{\partial z_i^{(j)}} . 
\end{equation}

In (2), we thus calculate the mean squared sensitivity across different N instances and take the root. The sensitivity for a particular instance (3) is obtained by first taking the derivative of an output node value w.r.t. a value in a patient representation; then taking the derivative of the patient representation value w.r.t. the original input value; and then multiplying them. This technique allows us to identify the most significant features in a trained model for an arbitrary number of instances and output classes. It is also transferable to the doc2vec representations, but we reserve this for future research.

\section{Dataset construction and preprocessing}
\label{sec:dataset}
We retrieve a set of adult patients ($\geq$18 years age) with only one hospital admission, with at least one associated textual note (excluding discharge reports) from the MIMIC-III critical care database \citep{johnson2016mimic}. We restrict to the patients with a single admission to remove ambiguity when the labels are dependent on discharge time. We exclude discharge reports from analyses to remove the direct indication of in-hospital death of a patient, which is one of the tasks that we are interested in. We obtain a range of 1--879 notes per patient, with average of 29.51 notes. This corresponds to 13--789,906 tokens per patient, with an average of 13,064 tokens. We split the dataset into 80-10-10\% as training, validation, and test subsets, to get a set of 24,650 patients for training, and 3,081 patients each for validation and testing. We represent patients with a concatenation of all the notes associated with them (excluding discharge reports). We tokenize the dataset using the Ucto tokenizer~\citep{van2012ucto} and lowercase it.

To obtain patient representations using the SDAE and for the baseline experiments, we replace the numbers, and certain token-level time and measurement matches with placeholders. We remove the punctuations, and the terms with corpus frequency less than 5. We represent the out-of-vocabulary terms obtained after the preprocessing in the test set with a common token. We use two feature sets---a bag-of-words (BoW), and a bag-of-medical-concepts (BoCUI)---with their corresponding TF-IDF scores 
as feature values. We use the TF-IDF values to give high weights to frequent features for a patient relative to all the patients in the dataset. For the BoCUI, we use the CLAMP toolkit~\citep{doi:110.1093/jamia/ocx132} to identify Concept Unique Identifiers (CUIs) in the UMLS\textsuperscript{\textregistered} metathesaurus~\citep{LindbergEtAl1993} corresponding to medical concept mentions of the types problems, treatments, and tests as defined in the i2b2 annotation guidelines~\citep{uzuner20112010}, along with their assertion labels. Here, problems also include findings and symptoms. CUIs appended with `present' and `absent' assertion labels are the vocabulary terms for this feature set. A bag-of-medical-concepts is a common featurization technique used in clinical NLP research~\citep{miotto2016deep,SCHEURWEGS2017}. We use a bag representation instead of a sequence model because the final document length for different patients is highly variable, going up to very large document sizes. We obtain a vocabulary size of 71,001 for the BoW feature set, and 83,310 for the BoCUI feature set.

To train the doc2vec models, we remove the numbers and the tokens matching certain time and measurement regex patterns. We have determined these settings based on the initial results on the validation set. We obtain a vocabulary size of 48,950 for this model. We have not trained a doc2vec model using only the medical concepts because if we represent a document as a sequence of CUIs only, we remove the indicators of language semantics from the context window, which the doc2vec model relies on during the learning process. If we keep additional terms along with the concept identifiers to train a doc2vec model, the available information is not comparable to a BoCUI feature set.

\section{Evaluation}

\subsection{Task description}
We use the dense patient representations as input features to train feedforward neural network classifiers on multiple independent tasks. We evaluate the performance on a range of tasks to gain insight into the task independent nature of the representations, and the information encoded within the vectors. We disregard the instances that do not have a task label. We minimize the categorical cross-entropy error using the RMSProp optimizer, and determine the hyperparameters using randomized search, which can be found in Table~\ref{hyperparams-ffnn} in the Appendix. 

\begin{enumerate}
\item \textbf{Patient mortality prediction:} Whether a patient dies within a given time frame. This prediction gives an estimate of the severity of a patient's condition to decide the amount of attention required.
\begin{enumerate}
\item \textbf{In-hospital mortality (In\_hosp):} Patient death during the hospital stay---13.14\% of the instances in the dataset.
\item \textbf{30 days mortality (30\_days):} Patient death within 30 days of discharge---3.85\% of the instances in the dataset.
\item \textbf{1 year mortality (1\_year):} Patient death within 365 days of discharge---12.19\% of the instances in the dataset. This includes the patients who died within 30 days of discharge.
\end{enumerate}

\item \textbf{Primary diagnostic category prediction (Pri\_diag\_cat):} Correctly diagnosing patients is essential for deciding further course of action. We evaluate if the proposed technique can be used to predict the generic category of the most relevant diagnostic code for a patient, corresponding to the 20 categories in the first volume of the 9th revision of the International Classification of Diseases, Clinical Modification (ICD-9-CM) database~\citep{world2004international}. A distribution of these categories in the dataset is given in Figure~\ref{fig:diag_proc_stats}. 

\begin{figure}

\begin{subfigure}{\textwidth}
\includegraphics[width=\linewidth]{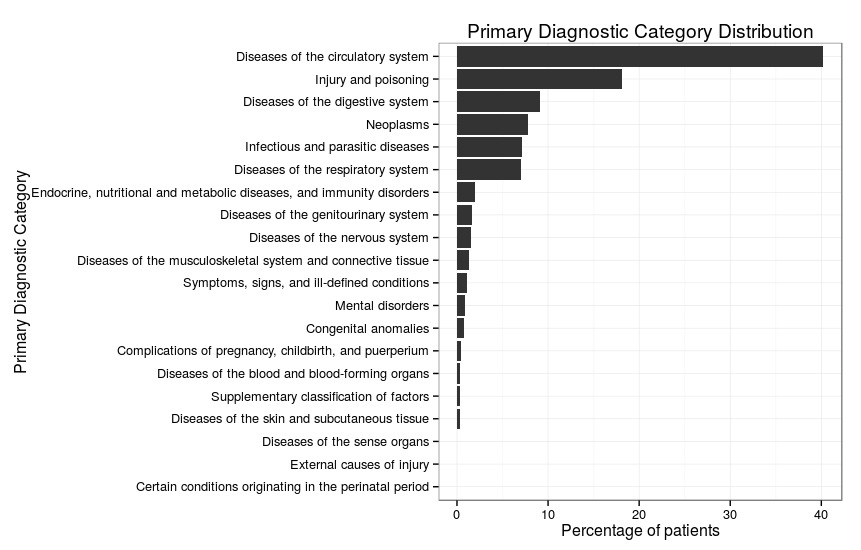}
\label{fig:diag_cat_stats}
\end{subfigure}
\begin{subfigure}{\textwidth}
\includegraphics[width=\linewidth]{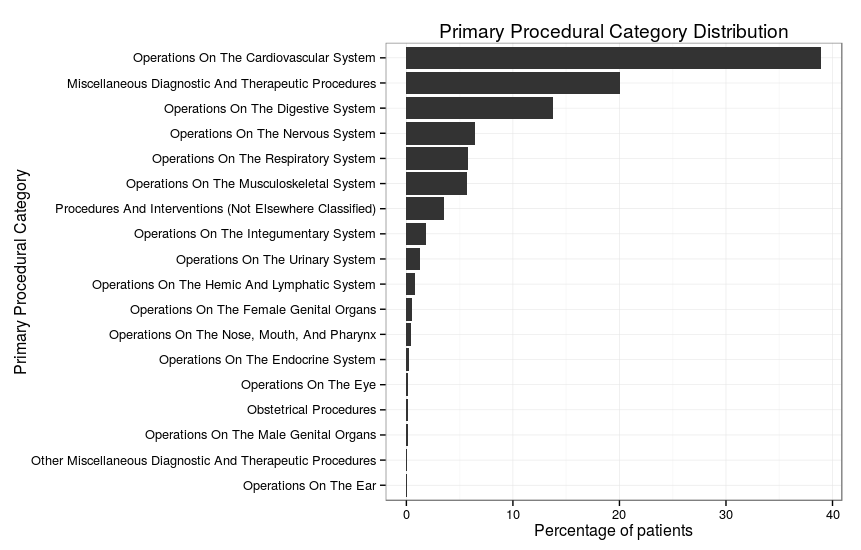}
\label{fig:proc_cat_stats}
\end{subfigure}

\caption{Primary diagnostic and procedural category distribution in the data.}
\label{fig:diag_proc_stats}
\end{figure}

\item \textbf{Primary procedural category prediction (Pri\_proc\_cat):} Predicting the generic category of the most relevant procedure performed on a patient, corresponding to the 18 categories present in the third volume of the ICD-9-CM database. A distribution of these categories in the dataset is given in Figure~\ref{fig:diag_proc_stats}. These procedural categories reflect different surgeries performed on patients. Prediction of the recommended procedure would assist the medical staff, while enabling optimal resource allocation for the same.

\item \textbf{Gender:} Gender of a patient---male (56.87\% of the instances) or female (43.13\% of the instances), as encoded in the dataset.
\end{enumerate}

\begin{figure}
\includegraphics[width=0.95\linewidth]{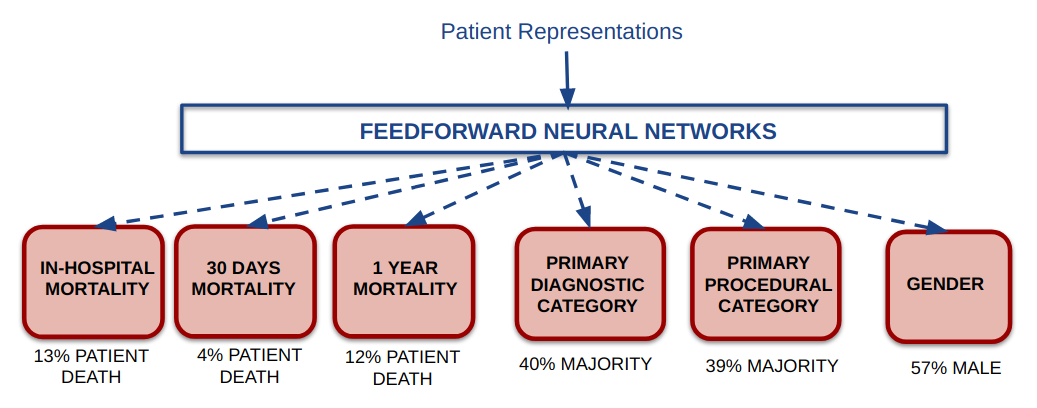}
\caption{Representation evaluation pipeline. The dashed lines indicate one of several operations, and are not performed in parallel.
}
\label{fig:eval}
\end{figure}

We evaluate the models using the area under the ROC curve (AUC-ROC) for patient death for the mortality tasks. The ROC curve gives us insight into the trade-off between the true positive rate and the false positive rate at different thresholds for different models. For the other tasks, we compute the weighted F-score to correct for class imbalance. We present the classification pipeline in Figure~\ref{fig:eval}.

\subsection{Results and Discussion}
\label{sec:results}

\subsubsection{Supervised Representation Evaluation}

\begin{table}[t]
  \caption{Classification results on different tasks using the BoW features, the SDAE representations computed from the BoW (SDAE-BoW), the doc2vec representations, the concatenated SDAE-BoW and doc2vec representations ([doc2vec, SDAE-BoW]) with Cohen's $\kappa$ score, the BoCUI features, and the SDAE vectors computed from the BoCUI (SDAE-BoCUI). AUC-ROC values are reported for the mortality tasks, and weighted F-score for the others.}
  \label{tab:results}
  \centering
  \begin{adjustbox}{max width=\textwidth}
  \begin{tabular}{llrrr|rrr}
    \toprule
    \multirow{2}{*}{\textbf{No.}} & \multirow{2}{*}{\textbf{Approach}} & \multicolumn{3}{c|}{\textbf{Mortality}} & \multirow{2}{*}{\textbf{Pri\_diag\_cat}} & \multirow{2}{*}{\textbf{Pri\_proc\_cat}} & \multirow{2}{*}{\textbf{Gender}} \\
  &  & \textbf{In\_hosp}  & \textbf{30\_days} &\textbf{ 1\_year} &  &  & \\
    \midrule
   (1) & BoW & 0.9457 & 0.5949 & 0.7942 & 0.7016 & 0.7366 & 0.9847\\
   (2) & SDAE-BoW  & 0.9194 & 0.7965 & 0.7980 & 0.6500 & 0.6746 & 0.8775 \\
   (3) & doc2vec  & 0.9195 & 0.7680 & 0.8134 & 0.6807 & 0.6583 & 0.9770 \\
   (4) & \makecell[l]{[doc2vec, SDAE-BoW]\\$\kappa$ \textit{score}} & \makecell[r]{0.9383\\\textit{0.5865}}  & \makecell[r]{0.8113\\\textit{0.0000}} & \makecell[r]{0.8302\\\textit{0.1581}} & \makecell[r]{0.6788\\\textit{0.6438}} & \makecell[r]{0.7030\\\textit{0.5891}} & \makecell[r]{0.9747\\\textit{0.7200}} \\
   (5) & BoCUI & 0.9088 & 0.5065 & 0.6993 & 0.7104 & 0.7265 & 0.7504 \\
   (6) & SDAE-BoCUI & 0.9007 & 0.7832 & 0.8016 & 0.6647 & 0.6777 & 0.6245 \\
    \bottomrule
  \end{tabular}
  \end{adjustbox}
\end{table}

\begin{figure}
\centering

\begin{subfigure}{0.7\textwidth}
\includegraphics[width=0.95\linewidth]{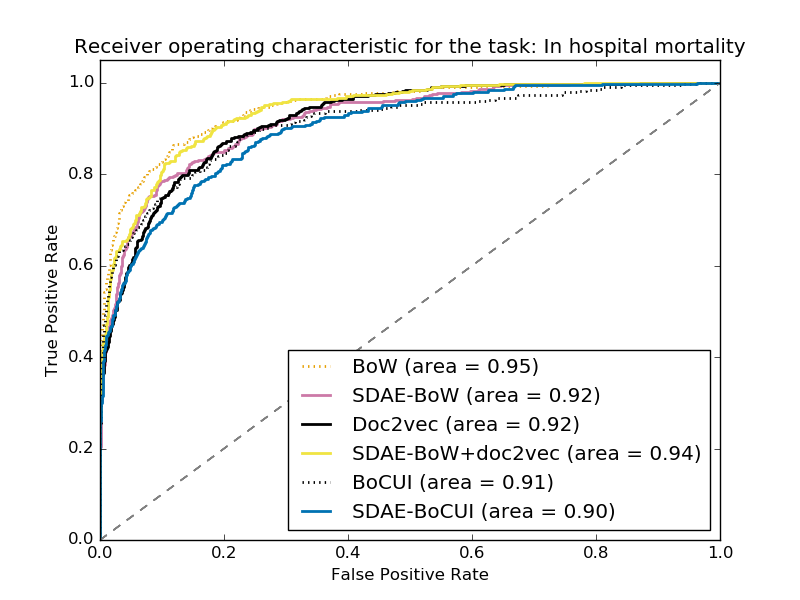}
\label{fig:roc_in_hosp}
\end{subfigure}
\begin{subfigure}{0.7\textwidth}
\includegraphics[width=0.95\linewidth]{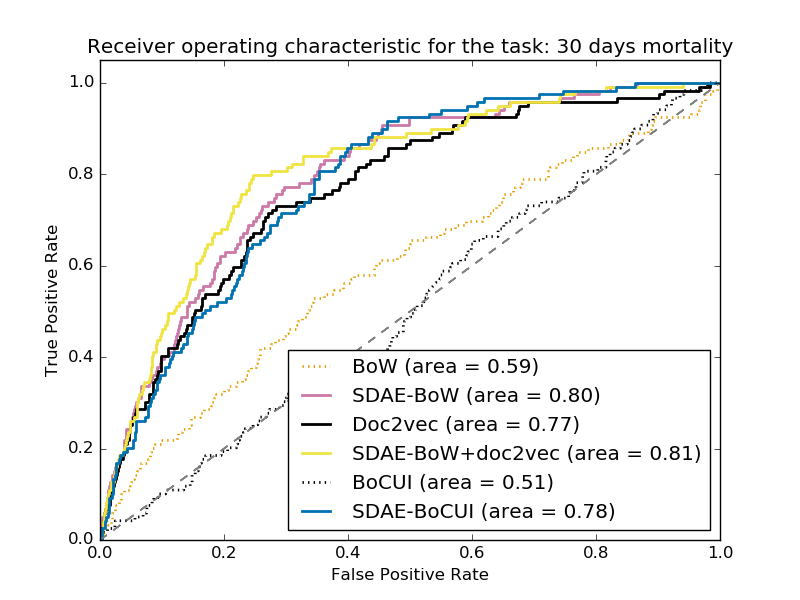}
\label{fig:roc_30_days}
\end{subfigure}
\begin{subfigure}{0.7\textwidth}
\includegraphics[width=0.95\linewidth]{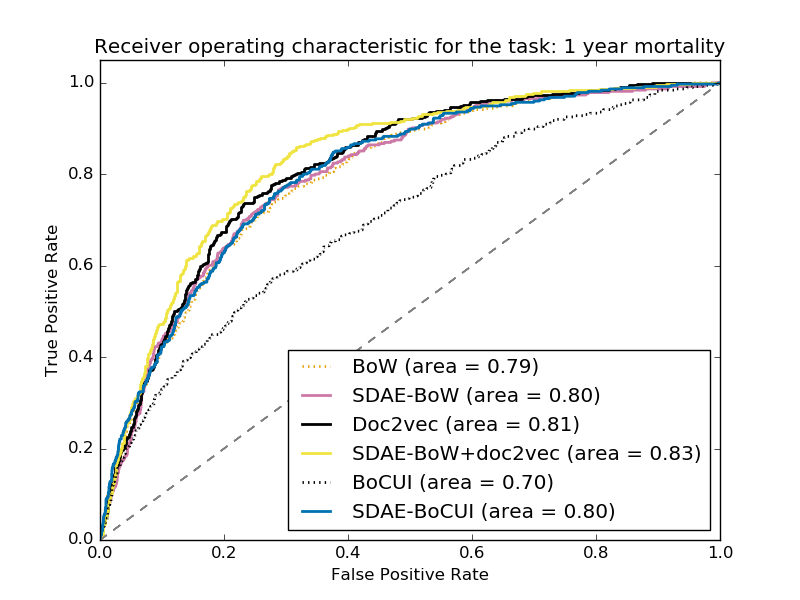}
\label{fig:roc_1_year}
\end{subfigure}

\caption{Receiver operating characteristic (ROC) for patient mortality prediction tasks.}
\label{fig:roc_curve}
\end{figure}

In Table~\ref{tab:results}, we compare the classification performance when we use the dense patient representations obtained from the SDAE-BoW (the initial SDAE input is BoW), the SDAE-BoCUI (the initial SDAE input is BoCUI), and the doc2vec models as input features for different tasks, as opposed to using the BoW and the BoCUI sparse features. In Figure~\ref{fig:roc_curve}, we show the ROC curves for the mortality prediction tasks. Further, we analyze the agreement between the SDAE-BoW and the doc2vec model outputs by calculating Cohen's $\kappa$ score~\citep{cohen1960coefficient} between them on the validation set. We find that the agreement scores are not high, which may indicate that the models learn complimentary information. We then concatenate the two dense representations (model ensemble) to analyze model complementarity. We calculate the statistical significance between the 9 different feature sets for the 6 tasks using the two-tailed pairwise approximate randomization test~\citep{noreen1989computer} with a significance level of 0.05 before the Bonferroni correction for 54 hypotheses\footnote{These hypotheses are the comparisons of the doc2vec, the SDAE-BoW, and the ensemble dense representations respectively with the BoW model, the ensemble with the doc2vec model, the ensemble with the SDAE-BoW model, the BoCUI with the BoW models, the SDAE-BoW model with the SDAE-BoCUI model, and the BoCUI model with the SDAE-BoCUI model for the 6 tasks.}.

Our main finding is that all the dense representation techniques significantly outperform the BoW baseline for 30 days mortality prediction. However, although we see a large numerical improvement over the BoW baseline on using the dense representations for 1 year mortality prediction (where the set of instances with the label `death' is a superset of those for 30 days mortality), the differences are not statistically significant. The SDAE-BoCUI model is significantly better than the BoCUI model for both 30 days and 1 year mortality prediction tasks. We believe that the poor performance of the sparse models for 30 days mortality prediction may be due to the low number of positive instances. The generalization afforded by the dense representation techniques assists feature identification in such cases. The sparse BoW inputs perform better than the SDAE-BoW representations for all the other tasks, and better than the doc2vec representations for in-hospital mortality and primary procedural category prediction. One probable reason is that the best predictors for the other tasks are the direct lexical mentions in the notes, which makes the BoW model a very strong baseline. Examples of such features obtained using the $\chi^2$ feature analysis are `autopsy', `expired', `funeral', and `unresponsive' for in-hospital mortality prediction, and `himself', `herself', `ovarian', and `testicular' for gender prediction. It is interesting to point out that the direct mentions of in-hospital death are present in the notes even though discharge reports have been excluded from analysis.

The agreement scores between the doc2vec and the SDAE-BoW models are not high for any task, which may indicate that the two models are complementary to each other. The results obtained from concatenation of the vectors learned by both models is not significantly different from the sparse representations for any task except 30 days mortality prediction, where the concatenation is better. This ensemble model significantly outperforms both individual models for primary procedural category prediction. For primary diagnostic category and gender prediction, the ensemble model is significantly better than the SDAE model, but not the doc2vec model. In these cases, there is no significant difference between the doc2vec and the BoW models. Hence, we observe that the concatenation helps in some cases and we recommend combining the two dense representations for unknown tasks. The doc2vec model uses a local context window in a log-linear classifier, whereas the SDAE model uses only the global context information and non-linear encoding layers. This may be one of the factors governing the differences between the two techniques. 

Furthermore, we observe that the BoCUI sparse features perform significantly worse than the BoW sparse features for in-hospital mortality, 1 year mortality, and gender prediction. For the other tasks, there is no statistical difference between the performance of the BoW and the BoCUI features, although we see a large numerical drop of about 9\% with the BoCUI model for 30 days mortality prediction. Moreover, the SDAE-BoW and SDAE-BoCUI representations are also not significantly different from each other for any of the tasks. These results suggest that there is no advantage of using a bag-of-concepts over a bag-of-words feature set, either as sparse inputs, or to learn dense representations. There are a few possible reasons behind the observed performance drop on using the BoCUI feature set. First, these features are restricted to the medical concepts of types `problem', `treatment', and `test'. These concepts are important features for diagnostic and procedural category identification. However, when we remove the terms that do not belong to these types, we also remove some useful features for other tasks, e.g., pronouns for gender prediction, and terms like `expired' and `post-mortem' for in-hospital mortality prediction, which in turn affects the classification performance. Next, when we identify medical concepts mentions with their corresponding CUIs and assertion labels, we also propagate the errors along in the pipeline, while adding to the sparsity of the terms. These factors additionally contribute to a difference in the classification performance.

Our work on mortality prediction is related to~\citet{grnarova2016neural}. The closest comparison between our results is the evaluation of the doc2vec representations. They have reported the AUC-ROC scores of 0.930, 0.831, and 0.824 for in-hospital mortality, 30 days mortality, and 1 year mortality prediction respectively, and have shown an improvement over the LDA baseline for the latter two. These scores are higher than what we have obtained with doc2vec. However, this may be due to different data subsets,\footnote{We were unable to reconstruct exact data subsets and obtain comparable results because we did not have access to their data processing scripts and the complete pipeline.} different classifiers (feedforward neural networks vs. linear SVMs), or different training schemes. They have further reported significant improvement on all the tasks when using a CNN architecture. This setup is supervised for the mortality tasks, and it is unclear whether supervision plays a role in the observed improvement. Similarly,~\citet{jocombining} have shown significant improvements for mortality prediction tasks on using their supervised LSTM architecture that jointly learns topic models as opposed to using LDA with linear SVMs. Again, the results are not directly comparable. They have predicted in-hospital, 30 days post-discharge, and 1 year post-discharge mortality at the end of every 12 hour window during a patient stay. Instead, we predict these mortality values using all the notes (except discharge reports) until the end of the patient stay. They have not reported the AUC-ROC scores for patient mortality at the end of the patient stay.

Furthermore,~\citet{dubois2017learning} have evaluated their embed-and-aggregate and RNN architectures for patient representation learning on multiple tasks. They have found that the RNN trained in a supervised manner for diagnostic code prediction outperforms the other architectures for predicting future diagnostic codes. However, when these representations are transferred to other tasks, this advantage is not visible. For mortality prediction (within the time period of the patient records) on large datasets, the bag-of-concepts and embed-and-aggregate methods performed equally well, and outperformed the RNN architectures. The RNN architecture performed poorly also for prediction of future patient admission, and had a comparable performance to embed-and-aggregate method for future ER visit prediction. One explanation for better RNN performance for future diagnostic code prediction is that the representations obtained from the RNN encode important information about patient diagnoses due to their supervised training on a similar task. This is not the case for the other tasks where there is no improvement.

\subsubsection{Feature analysis}

\begin{table}[t]
\parbox{.45\linewidth} {
  \caption{The best and the worst feature reconstructions during unsupervised pretraining of SDAE-BoW.}
  \label{feat_recon_error}
  \begin{adjustbox}{max width=0.45\textwidth}
  \begin{tabular}{ll}
    \toprule
     Best reconstruction & Worst reconstruction \\
    \midrule
     stumnz & picc \\
     jajhnx & woman \\
     a-fibril & osh \\
     lsc.o & fall \\
     potentiallly & man \\
     yesh & stent \\
     forcal & he \\
     contbributing & wife \\
     hyponatremia-on & repair \\
     pre-exiusting & bleed \\
    \bottomrule
  \end{tabular}
  \end{adjustbox}
  }
  \hfill
\parbox{.45\linewidth}{
\caption{Correlation between the mean squared reconstruction error of the first layer of the SDAE during the unsupervised pretraining phase and feature frequency. All the p-values are lower than 0.001.}
  \label{freq_corr}
  \begin{adjustbox}{max width = 0.45\textwidth}
  \begin{tabular}{lrr}
    \toprule
     Feature set  & Spearman  & Kendall-Tau \\
    \midrule
    BoW & 0.8738 & 0.7287     \\
    BoCUI & 0.8836 & 0.7334   \\
    \bottomrule
  \end{tabular}
  \end{adjustbox}
}

\end{table}

In Table~\ref{feat_recon_error}, we present a list of features based on their mean squared reconstruction error when we pretrain the patient representations using the SDAE-BoW model. We observe that infrequent terms such as spelling errors are reconstructed very well, as opposed to the frequent features in the dataset. To check for a correlation between the mean squared reconstruction error and the feature frequency, we calculate the Spearman's and the Kendall-tau rank-order correlation coefficients~\citep{kokoska2000crc} between the two parameters, reported in Table~\ref{freq_corr}. These techniques check for a correlation between the parameters irrespective of a linear relationship and use different algorithms to generate the ranked lists in case of a tie. Using both techniques, we obtain very high positive correlation coefficients. We believe that this behavior may be either due to the high entropy of the frequent terms, or because the model memorizes the infrequent terms. \citet{jocombining} also obtain misspellings and rare words as the top features when they use recurrent neural networks for patient mortality prediction in the MIMIC-III dataset.

\begin{table}[t]

  \caption{The most significant features in ranked order for the classifiers for one instance each when the SDAE-BoW representations are used as the input. The true classes are `patient death' for the mortality tasks (a common instance for 30 days and 1 year mortality prediction), and `diseases of the digestive system', `operations on the digestive system', and `male' respectively for a common patient for the other tasks.}
  \label{feat_imp}
  \centering
  \resizebox{\textwidth}{!}{
  \begin{tabular}{llllll}
    \toprule
     In\_hosp & 30\_days & 1\_year & Pri\_diag\_cat & Pri\_proc\_cat & Gender \\
    \midrule
     \textit{vasopressin} & leaflet & \textit{magnevist} & \textit{numeric\_val} & \textit{numeric\_val} & \textit{woman} \\
     pressors & \textit{structurally} & \textit{signal} & previous & no & \textit{female} \\
     \textit{focused} & \textit{pacemaker} & \textit{decisions} & rhythm & of & \textit{she}\\
     \textit{dnr} & \textit{sda} & \textit{periventricular} & no & \textit{enzymes} & man \\
     dopamine & \textit{periventricular} & \textit{embolus} & \textit{flexure} & \textit{extubated} & he \\
     acidosis & \textit{excursion} & \textit{underestimated} & \textit{dementia} & rhythm & male\\
     levophed & \textit{non-coronary} & \textit{calcified} & brbpr  & and & \textit{her} \\
     pressor & \textit{dosages} & \textit{screws} & of  & the & his \\
     \textit{cvvhd} & \textit{microvascular} & \textit{rib} & sinus & \textit{vent} & wife \\
     \textit{cvvh} & left-sided & \textit{shadowing} & for & \textit{uncal} & \textit{uterus} \\
     \textit{emergency} & \textit{chronic} & \textit{gadolinium} & to & \textit{mso} & him \\
     \textit{pneumatosis} & \textit{extubation} & \textit{mri} & tracing & to & \textit{urinal} \\
    \bottomrule
  \end{tabular}}
\end{table}

In Table~\ref{feat_imp}, we list the most significant features for the model output for one instance each in the test set, when the SDAE-BoW patient representations $R$ are used as the classification input. In italics are the vocabulary terms that are not present in the notes for the patient, but are treated as the most influential features. We find that the classifiers give high importance to sensible frequent features for most of the tasks, although the SDAE reconstructs low frequent terms such as spelling errors better during the pretraining phase. Several features for in-hospital mortality point towards the overall patient condition and treatments for the patient. Terms like `brbpr' (bright red blood per rectum) for primary diagnostic category prediction, and the top features for gender prediction indicate the true class. The absence of several features is used as an important clue to identify the right class. For example, most of the top ranking features for 30 days and 1 year mortality prediction are not present in the patient notes. Similarly, the absence of the terms related to the female gender implies the male class. Additionally, the absence of numbers (`numeric\_val') in notes is the most useful feature for diagnostic and procedural category identification, which may have been used by the model to identify certain lab tests with numeric results that were not carried out. 

Furthermore, many top features extracted for primary diagnostic category prediction are the terms corresponding to text segments like \textit{``Sinus rhythm. Compared to the previous tracing of ..."}, which is a common pattern in the notes for the patient. When evaluated without the context, many of these terms do not make sense. However, although we input a bag-of-words representation to the SDAE, co-occurrence of the terms is reflected in the extracted features. We further observe that there is a minimal overlap between the sets of important features for different tasks. This shows that the learned representations $R$ are task-independent, and that the classifiers can identify task-specific important information when they are trained for a particular task.

\begin{table}[t]

  \caption{Comparison of the best features for one instance of in-hospital patient death, where the BoW model makes the correct prediction and the SDAE-BoW model fails, and for one instance where both the models make the correct prediction.}
  \label{error_analysis}
  \centering
  \begin{tabular}{ll|ll}
    \toprule
     BoW  & SDAE-BoW  &  BoW  & SDAE-BoW \\
    (correct) & (incorrect) &  (correct) & (correct)\\
    \midrule
     expired & cad  & expired & vasopressin \\
     autopsy & cabg & autopsy & pressors\\
     cmo & pre-op & morgue & focused\\
     pre-bypass & preop & cmo & dnr\\
     morgue & numeric\_val & toradol & dopamine\\
     diseasecoronary & no & diseasecoronary & acidosis\\
     deline & bypass  & deline & levophed\\
     prebypass & sternotomy & prebypass & pressor\\
     death & lat & pre-bypass & cvvhd\\
     decannulation & ptx & asystolic & cvvh\\
    \bottomrule
  \end{tabular}
\end{table}

To illustrate the applicability of the feature extraction technique to understand relative model behavior, we compare the set of the most important features for a) one instance where the bag-of-words model predicts in-hospital death correctly, whereas the SDAE dense representations fail to make that prediction, and b) one instance where both the models make correct predictions. These features are presented in Table~\ref{error_analysis}. We find that the BoW model identifies the direct indicators of patient death such as `expired', `autopsy', `morgue', and `death' as the top features along with certain features related to the procedures performed on the patient. Instead, the generalized SDAE-BoW model uses the features related to the holistic patient condition as the more important features. Examples are `cad (Coronary Artery Disease)', `cabg (Coronary Artery Bypass Graft surgery)', `vasopressin', `dopamine', `dnr (do not resuscitate)', and `cvvhd (Continuous Veno-Venous Hemofiltration Dialysis)'. This shows us that the models operate in very different feature spaces. The generalized models are good when we want a comprehensive view of the patient condition. However, the sparse BoW model may be better if we want to pick up the strong lexical features present for a task.

\subsubsection{Visualization of Unsupervised Representations}

\begin{figure}

\centering

\begin{subfigure}{\textwidth}
\includegraphics[width=\linewidth]{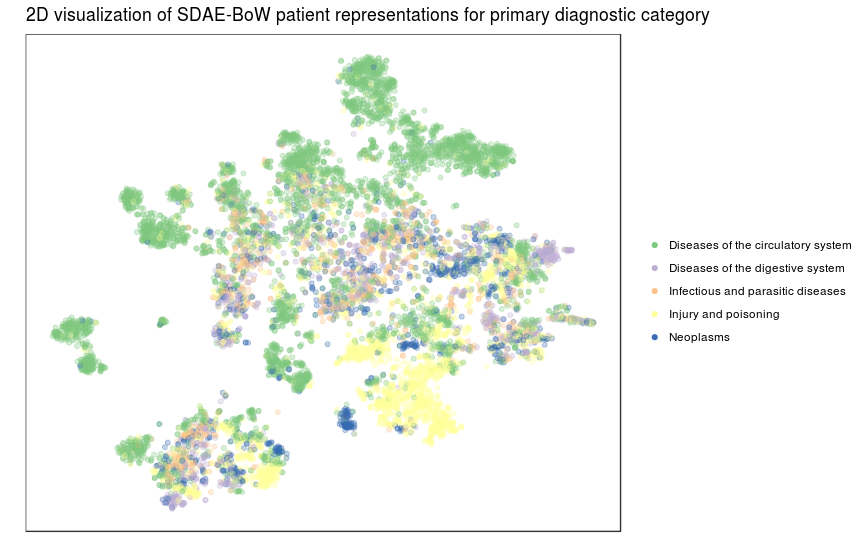}
\end{subfigure}

\begin{subfigure}{\textwidth}
\includegraphics[width=\linewidth]{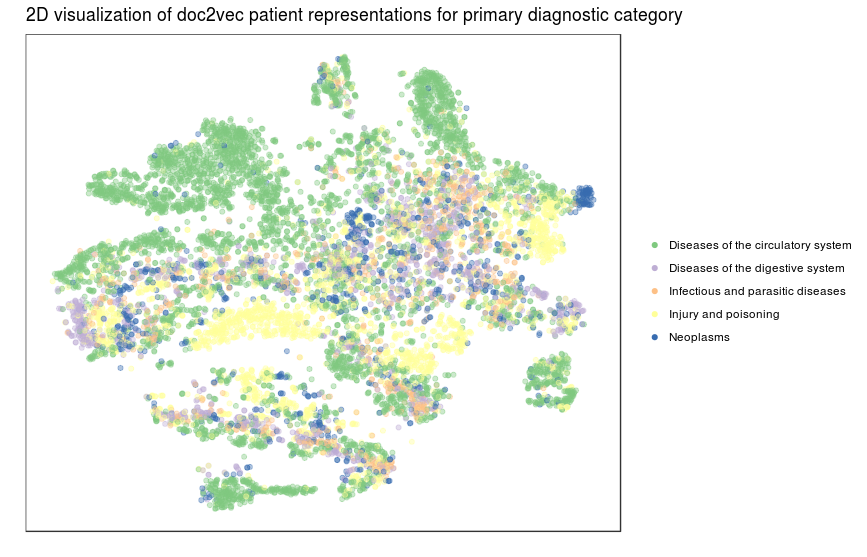}
\end{subfigure}

\caption{t-SNE visualization of SDAE-BoW and doc2vec representations.}
\label{fig:tsne}
\end{figure}

In Figure~\ref{fig:tsne}, we present 2D visualizations of the unsupervised representations learned by the SDAE and the doc2vec architectures. It is important to note that the SDAE-BoW and the doc2vec representations were learned in an unsupervised manner, and were not finetuned to represent a particular property of the data. Hence, they encode information that represent patient notes in a holistic manner, and span many different properties. We use t-SNE\footnote{We experimented with different values of perplexity and the number of iterations for the t-SNE. After converging at 5000 iterations, the resulting visualizations were similar across most perplexity values, albeit often rotated. We chose a perplexity of 50 for the SDAE-BoW representations, and 30 for the doc2vec representations.}~\citep{maaten2008visualizing} to generate the visualizations, after first reducing the representations to 50 dimensions\footnote{Nearly 70\% of the variation was explained by these 50 dimensions.} using Principal Component Analysis. In the figure, as an example, we color the representations according to the corresponding primary diagnostic category. For the purpose of clarity, we limit to the 5 most frequent diagnostic categories in the dataset. We observe that the patients with the same diagnostic category are frequently close together, forming clusters. This suggests that using the proposed techniques, ``similar'' patients result in similar representations.

\section{Conclusions and Future Work}

Our research provides insight into the suitability of learning patient representations only from clinical notes, for an arbitrary task, while understanding model performance. We have shown that the generalized dense patient representations significantly improve the classification performance for 30 days mortality prediction, a task where we are confronted with a very low proportion of positive instances. For the other tasks, this advantage is not visible. Moreover, we have shown that a combination of the stacked denoising autoencoder and the doc2vec representations improves over the individual models for some tasks, without any harm to the others tasks. We recommend combining these representations for unknown tasks. We have further shown that there is no advantage of using a bag-of-concepts feature set as opposed to a bag-of-words feature set as either sparse inputs or to learn dense representations. Expensive concept identification process is not required for these setups. 

Furthermore, we have proposed novel techniques to interpret model performance to overcome the black-box nature of neural networks. During representation analysis, we have found that frequent terms are not encoded well during the pretraining phase of the stacked denoising autoencoder. However, when we use these pretrained vectors as the input, sensible frequent features are selected as the most significant features for the classification tasks. Some vocabulary items that are absent from patient notes are often deemed important, while at the same time, co-occurrence of the features present in the notes is also learned by the model. We have also shown that the unsupervised representations are task-independent and distinct features are extracted for different tasks when these representations are used as supervised inputs.

This work lays down the path for more applied research in the clinical domain. In future, we plan to compute patient similarity from the generalized patient representations to identify patient cohorts. We also plan to add structured information to analyze their comparative contribution to the learned representations for the different tasks. Furthermore, the techniques that we have proposed to understand the behavior of statistical models are transferable to different architectures and facilitate further research in this crucial direction.

\section{Acknowledgements}
This research was carried out within the Accumulate VLAIO SBO project, funded by the government agency Flanders Innovation \& Entrepreneurship (VLAIO) [grant number 150056]. We thank Riccardo Miotto, Paulina Grnarova, and Florian Schmidt for sharing their model implementation details from related papers, and answering several questions about the same. Further, we thank Giovanni Cassani, St{\'e}phan Tulkens, and Ella Roelant for their help with statistical significance analyses. We would also like to thank the anonymous reviewers for their useful comments.

\section*{References}

\bibliography{mybibfile}

\appendix
\section{Model Hyperparameters}

\setcounter{table}{0}
\begin{table}[H]
  \caption{Hyperparameters for stacked denoising autoencoder to learn dense patient representations, obtained after a randomized search. The default learning rate of 0.001 is used.}
  \label{hyperparams-sdae}
  \centering
  \begin{tabular}{lrrr}
    \toprule
     Feature set & Number of layers & Hidden dimensions & Dropout proportion \\
    \midrule
     Bag-of-words & 1  & 800 & 0.05 \\
     Bag-of-concepts & 1 & 300 & 0.4\\
    \bottomrule
  \end{tabular}
\end{table}

\begin{table}[H]
  \caption{Hyperparameters for feedforward neural network classifiers for different tasks and feature sets, obtained after a randomized search. The default learning rate of 0.001 is used.}
  \label{hyperparams-ffnn}
  \centering
  \resizebox{\textwidth}{!}{
  \begin{tabular}{llrrl}
    \toprule
     Task & Feature set & Number of layers & Hidden dimensions & Activation function \\
    \midrule
     \multirow{6}{*}{In\_hosp}
     & BoW & 7 & 980 & sigmoid \\
     & SDAE-BoW & 7 & 160 & relu \\
     & doc2vec & 10 & 410 & sigmoid \\
     & [doc2vec, SDAE-BoW] & 7 & 340 & tanh \\
     & BoCUI & 3 & 680 & sigmoid \\
     & SDAE-BoCUI & 3 & 560 & sigmoid \\
     
     \midrule
     
     \multirow{6}{*}{30\_days} 
     & BoW & 10 & 220 & relu \\
     & SDAE-BoW & 3 & 820 & sigmoid \\
     & doc2vec & 2 & 900 & sigmoid \\
     & [doc2vec, SDAE-BoW] & 8 & 430 & sigmoid \\
     & BoCUI & 7 & 510 & tanh \\
     & SDAE-BoCUI & 3 & 750 & sigmoid \\
     
     \midrule
     
     \multirow{6}{*}{1\_year} 
     & BoW & 1 & 650 & sigmoid \\
     & SDAE-BoW & 10 & 570 & sigmoid \\
     & doc2vec & 3 & 1000 & sigmoid \\
     & [doc2vec, SDAE-BoW] & 5 & 920 & sigmoid \\
     & BoCUI & 1 & 290 & sigmoid \\
     & SDAE-BoCUI & 6 & 290 & relu \\
     
     \midrule
     
     \multirow{6}{*}{Pri\_diag\_cat} 
     & BoW & 4 & 100 & sigmoid \\
     & SDAE-BoW & 2 & 110 & sigmoid \\
     & doc2vec & 9 & 600 & relu \\
     & [doc2vec, SDAE-BoW] & 8 & 700 & relu \\
     & BoCUI & 4 & 80 & sigmoid \\
     & SDAE-BoCUI & 8 & 230 & relu \\
     
     \midrule
     
     \multirow{6}{*}{Pri\_proc\_cat} 
     & BoW & 2  & 220 & sigmoid \\
     & SDAE-BoW & 5 & 890 & relu \\
     & doc2vec & 3 & 980 & relu \\
     & [doc2vec, SDAE-BoW] & 8 & 520 & relu \\
     & BoCUI & 10 & 760 & relu \\
     & SDAE-BoCUI & 6 & 540 & relu \\
     
     \midrule
     
     \multirow{6}{*}{Gender} 
     & BoW & 0 & NA & NA \\
     & SDAE-BoW & 8 & 160 & relu \\
     & doc2vec & 0 & NA & NA \\
     & [doc2vec, SDAE-BoW] & 7 & 280 & sigmoid \\
     & BoCUI & 5 & 410 & relu \\
     & SDAE-BoCUI & 1 & 210 & relu \\
    \bottomrule
  \end{tabular}}
\end{table}
\end{document}